\documentclass[journal]{IEEEtran}

\ifCLASSINFOpdf
\else
\fi
\usepackage{graphicx}
\usepackage{amsmath}
\usepackage{cite}
 \newcommand{\etal}{\textit{et al}.}
\newcommand{\ie}{\textit{i}.\textit{e}.}
\newcommand{\eg}{\textit{e}.\textit{g}.}
\usepackage[colorlinks,linkcolor=red]{hyperref}

\begin{document}

\title{Emerging from Water: Underwater Image Color Correction  Based on Weakly Supervised Color Transfer}

\author{Chongyi~Li,~\IEEEmembership{Student~Member,~IEEE,}
        Jichang~Guo,
        Chunle~Guo

\thanks{This work was supported in part by the National Key Basic Research Program of China (2014CB340403), the National Natural Science Foundation of China (61771334).}

\thanks{Chongyi Li, Jichang Guo and Chunle Guo are with the School of Electrical and Information Engineering, Tianjin University, Tianjin, China (e-mail: lichongyi@tju.edu.cn; jcguo@tju.edu.cn; guochunle@tju.edu.cn).

(Corresponding author: Chunle Guo.)}}
\markboth{IEEE single processing letters}%
{Shell \MakeLowercase{\textit{et al.}}: Bare Demo of IEEEtran.cls for Journals}

\maketitle

\begin{abstract}

Underwater vision suffers from severe effects due to selective attenuation and scattering when light propagates through water. Such degradation not only affects the quality of underwater images but limits the ability of vision tasks. Different from existing methods which either ignore the wavelength dependency of the attenuation or assume a specific spectral profile, we tackle color distortion problem of underwater image from a new view. In this letter, we propose a weakly supervised color transfer method to correct color distortion, which relaxes the need of paired underwater images for training and allows for the underwater images unknown where were taken. Inspired by Cycle-Consistent Adversarial Networks, we design a multi-term loss function including adversarial loss, cycle consistency loss, and SSIM (Structural Similarity Index Measure) loss, which allows the
content and structure of the corrected result the same as the input, but the color as if the image was taken without the water. Experiments on underwater images captured under diverse scenes show that our method produces visually pleasing results, even outperforms the art-of-the-state methods. Besides, our method can improve the performance of vision tasks.

\end{abstract}

\section{Introduction}

Recently, ocean engineering and research have increasingly relied on underwater images captured from Autonomous Underwater Vehicles (AUVs) and Remotely Operated Vehicles (ROVs) \cite{Bryson2016}. However, underwater images usually suffer from degeneration, such as low contrast, color casts, and noise, due to wavelength-dependent light absorption and scattering as well as the effects from low-end optical imaging devices. The scattering and absorbtion attenuate the direct transmission and introduce surrounding scattered light. The attenuated direct transmission causes the intensity from the scene to weaker and color casts, while the surrounding scattered light causes the appearance of the scene to be washed out. Besides, the magnitude of attenuation and scattering depends on several complex factors including water temperature and salinity, and the type and quantity of particulates in the water. Serious degeneration makes it difficult to recover the appearance and color of underwater images. However, color is extremely important for underwater vision tasks and research \cite{Gibson2016}. Therefore, how to effectively approximate the real color of underwater image has become a challenging problem needed to be solved.

A number of methods \cite{He2004, Schechner2004, Schechner2005, Treibitz2009, Carlevaris2010, Chiang2012,Galdran2015, Drews2016, Li2016, Peng2017, Li2017, Zhang2017, Ahn2017,Emberton2017} have been proposed to improve the visual quality of underwater images,
ranging from hardware solutions to image dehazing and color correction methods \cite{Schettini2010, Haware2017}.
The hardware solutions \cite{He2004, Schechner2004, Schechner2005, Treibitz2009} have shown the effectiveness,
but these solutions are not applicable to dynamic acquisition. Most of single underwater image restoration
methods \cite{Carlevaris2010, Chiang2012,Galdran2015, Drews2016, Li2016, Peng2017, Emberton2017} are inspired by the outdoor dehazing strategies \cite{He2011, Zhu2015, Tang2014, Cai2016}. For underwater image color
correction, traditional color constancy methods (\eg, Gray World \cite{Buchsbaum1980}, Max RGB \cite{Land1977},
White Balance \cite{Ebner2007}, Shades-of-Grey \cite{Finlayson2004}, and \etal) and their variations are usually employed. Compared with traditional methods based on statistical priors, we bridge the gap between the color of underwater image and that of air image by learning their cross domain relations. Recently, a semi-supervised learning model for underwater image color correction, namely WaterGAN, has been proposed \cite{Li2017}. Unlike this work, our model is no need for a large annotated dataset of images pairs. Besides, WaterGAN shows limitation in processing the underwater images captured under unknown sites. Contrast to translating image style \cite{Goodfellow2014, Radford2016}, our model learns the semantic color of air image and preserves the key attributes such as content and structure of input image.

\begin{figure}[tp]
\begin{center}
\begin{tabular}{c@{ }  c@{ } c@{ }  c@{ } }
\includegraphics[width=.11\textwidth]{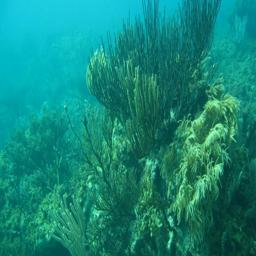} &
\includegraphics[width=.11\textwidth]{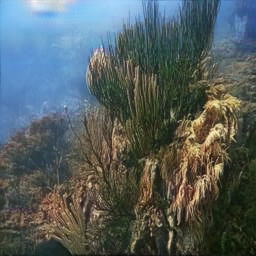} &
\includegraphics[width=.11\textwidth]{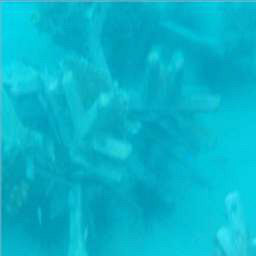} &
\includegraphics[width=.11\textwidth]{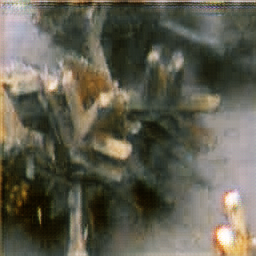}\\
(a) & (b) & (c) & (d)\\
\end{tabular}
\end{center}
\vspace{-2mm}
\caption{Two examples of our results. (a) and (c) are raw underwater images. (b) and (d) are our results. }
\label{fig:example}
\vspace{-2mm}
\end{figure}

In this letter, we present a novel weakly supervised model for underwater image color correction, which maps the color from the scenes of underwater into the scenes of air without any explicit pair labels. Specifically, given an underwater image as the input, our model directly outputs an image, which has the content and structure the same as the input, but the color as if the image was taken without the water. Though the translated color might not be the ``real'' color of underwater image, our method removes the color casts and improves the performance of vision tasks. In fact, it is impossible that a method recovers the appearance and color of any underwater images when the scenes and light conditions are unknown. Figure~\ref{fig:example} shows two examples of our attempts.

\textbf{Contributions.} This letter introduces the following main contributions: \textbf{(1)} Compared to existing priors/assumptions based and semi-supervised methods,  to our best knowledge, this is the first attempt to build a weakly supervised model for underwater image color correction. Here, ``weak supervision'' means that our model relaxes the need of paired underwater images for training and allows for the underwater images unknown where were taken. \textbf{(2)} We tackle underwater image color correction problem from a new angle, which learns a cross domain mapping function between underwater images and air images.  \textbf{(3)} A multi-term loss function allows our model capturing context and semantic information, which preserves the content and structure of input image, but the color as if the output image was taken without the water.

\section{Proposed Method}
\begin{figure*}[tp]
\begin{center}
\begin{tabular}{c@{ }  c@{ } }
\includegraphics[width=.51\textwidth]{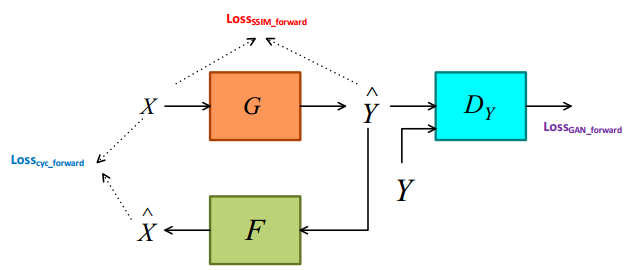}&
\includegraphics[width=.51\textwidth]{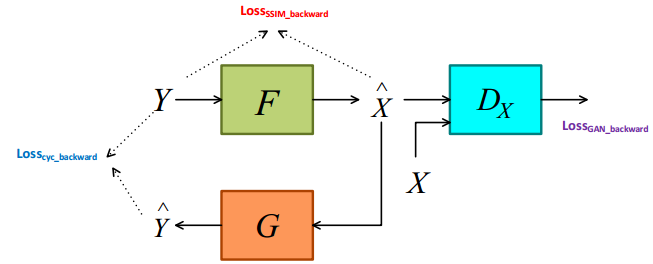}\\
Forward networks& Backward networks\\
\end{tabular}
\end{center}
\vspace{-2mm}
\caption{A diagram of our framework. Our model contains two mapping functions $G: X\rightarrow Y$ (forward) and $F: Y \rightarrow X$ (backward), and associated adversarial discriminators $D_{Y}$ and $D_{X}$.}
\label{fig:framework}
\vspace{-2mm}
\end{figure*}

Our solution is based upon very recent advances in image-to-image translation networks \cite{Zhu2017, Kim2017, Liu2017}, which captures special characteristics of one image collection and figures out how these characteristics could be translated into the other image collection, all in the absence of any paired training examples. Our goal is to learn mapping functions between a source domain $X$ (\ie, underwater) and a target domain $Y$ (\ie, air). The inputs are unpaired training image samples $x\in X$ and $y\in Y$. Specifically, our model includes two mappings $G: X\rightarrow Y$ (forward) and $F: Y\rightarrow X$ (backward) given training data $\{{x_{i}}\}^{N}_{i=1}$ $\in X$ and $\{{y_{i}}\}^{N}_{i=1}$ $\in Y$. Following the CycleGAN \cite{Zhu2017}, we also use two adversarial discriminators $D_{X}$ and $D_{Y}$. The $D_{X}$ aims to distinguish between images $\{x\}$ and translated images $\{F(y)\}$ when the $D_{Y}$ aims to distinguish between images $\{y\}$ and translated images $\{G(x)\}$. For the generator of forward networks, the loss function includes three terms: adversarial loss is to match the distribution of generated images to the distribution in the target domain; cycle consistency loss is to prevent the learned mappings $G$ and $F$ from contradiction each other; SSIM loss \cite{Wang2004} is to preserve the content and structure of source images. For the generator of backward networks, the loss function also includes three terms. The framework of the proposed method is shown in Figure~\ref{fig:framework}. Below, we just present the losses of forward networks and vice versa for backward networks.

\subsection{Adversarial loss}
For the mapping function $G: X\rightarrow Y$ and its discriminator $D_{Y}$, the adversarial loss can be expressed as:
\begin{equation}
\label{equ_1}
\begin{aligned}
L_{GAN}(G,D_{Y},X,Y)&=E_{y\sim p_{data}(y)}[log D_{Y}(y)]\\
&+E_{x\sim p_{data}(x)}[log(1-D_{Y}(G(x))].
\end{aligned}
\end{equation}
where $G$ tries to generate images $G(x)$ that look similar to images from domain $Y$ while $D_{Y}$ aims to distinguish between translated samples $G(x)$ and real samples $y$.

\subsection{Cycle consistency loss}
We add a cycle consistency loss to constrain the space of possible mapping functions. Cycle consistency means that for each image $x$ from domain $X$, the image translation cycle should be able to bring $x$ back to the original image, while for each image $y$ from domain $Y$, the image translation cycle should be able to bring $y$ back to target image. Cycle consistency loss can be expressed as:
\begin{equation}
\label{equ_2}
\begin{aligned}
L_{cyc}(G,F)=&E_{x\sim p_{data}(x)}[\|F(G(x))-x)\|_{1}] \\
&+E_{y\sim p_{data}(y)}[\|G(F(y))-y\|_{1}].
\end{aligned}
\end{equation}
\subsection{SSIM loss}
To preserve the content and structure of the input image, SSIM loss \cite{Zhao2015} is used. First, for pixel $p$, the SSIM between input image $x$ and translated image $G(x)$ is defined as:
\begin{equation}
\label{equ_3}
SSIM(p)=\frac{2\mu_{x}\mu_{y}+C_{1}}{\mu_{x}^{2}+\mu_{y}^{2}+C_{1}}\cdot\frac{2\sigma_{xy}+C_{2}}{\sigma_{x}^{2}+\sigma_{y}^{2}+C_{2}},
\end{equation}
where $p$ is the center pixel of an image patch, $x$ is an image patch $\in X$ with size $13\times13$, $y$ is an image patch $\in G(x)$ with size $13\times13$, $\mu_{x}$ is the mean of $x$, $\sigma_{x}$ is the standard deviations of $x$, $\mu_{y}$ is the mean of $y$, $\sigma_{y}$ is the standard deviations of $y$,  $\sigma_{xy}$ is the covariance of $x$ and $y$. $C_{1}$=0.02 and $C_{2}$=0.03 are default in SSIM loss. In this way, the SSIM value for every pixel between input image $x$ and translated image $G(x)$ is calculated. The SSIM loss can be expressed as:
\begin{equation}
\label{equ_4}
L_{SSIM}(x,G(x))=1-\frac{1}{N}\Sigma_{p=1}^{N}(SSIM(p)).
\end{equation}
where $N$ is the number of pixel in an image.

\subsection{Total loss}
Finally, the total loss is the linear combination of the above-mentioned three losses with weights:
\begin{equation}
\label{equ_5}
\begin{aligned}
L_{loss}=&\lambda_{1}L_{GAN}(G,D_{Y},X,Y)+\lambda_{2}L_{cyc}(G,F)\\
&+\lambda_{3}L_{SSIM}(x,G(x)).
\end{aligned}
\end{equation}
where weights $\lambda_{1}$, $\lambda_{2}$ and $\lambda_{3}$ are 1, 1 and 10 based on heuristic experiments on our training data, which makes the order of magnitude of three components equal because we expect that these three components have equally significant contributions in the final loss function. The optimization of generator $G$ of forward networks is to minimize Equation~\eqref{equ_5} when the optimization of discriminator $D_{Y}$ is to maximize Equation~\eqref{equ_1}. In training process, we alternatively optimize $G$, $D_{Y}$, $F$, and $D_{x}$, respectively.

\subsection{Network architecture and training details}
\begin{figure}[tp]
\begin{center}
\begin{tabular}{c@{ }  c@{ } c@{ }  c@{ } }
\includegraphics[width=.11\textwidth]{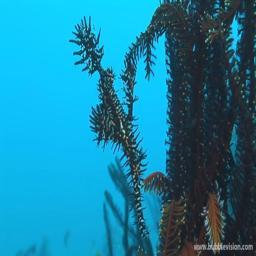}&
\includegraphics[width=.11\textwidth]{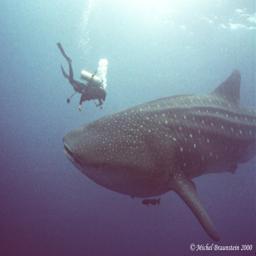}&
\includegraphics[width=.11\textwidth]{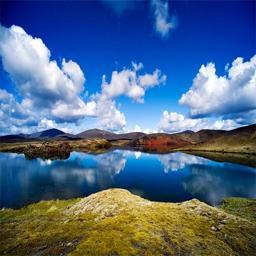} &
\includegraphics[width=.11\textwidth]{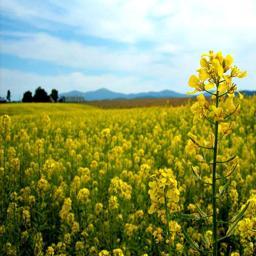}\\
(a) & (b) & (c) & (d)
 \end{tabular}
\end{center}
\vspace{-2mm}
\caption{Our training samples. (a) and (b) are underwater images. (c) and (d) are air images.}
\label{fig:samples}
\vspace{-2mm}
\end{figure}

The forward and backward networks have the same architecture. Following CycleGAN \cite{Zhu2017}, we adapt the architecture \cite{Johnson2016} for our generators and use $70\times70$ PatchGANs \cite{Isola2016, Ledig2016, Li2016eccv} as our discriminators. To train our model, we collect a dataset from Internet, which contains 3800 underwater images and 3800 air images. After that, those images are resized to $256\times256$ based on our limited memory and the architecture of our discriminators. Figure~\ref{fig:samples} shows several training samples.

In addition, to stabilize our model training, we replace the negative log likelihood objective by a least square loss. Equation~\eqref{equ_1} is rewritten as :
\begin{equation}
\label{equ_6}
\begin{aligned}
L_{GAN}(G,D_{Y},X,Y)&=E_{y\sim p_{data}(y)}[(D_{Y}(y)-1)^{2}]\\
&+E_{x\sim p_{data}(x)}[D_{Y}(G(x))^2].
\end{aligned}
\end{equation}

The discriminators $D_{X}$ and $D_{Y}$ were updated by a history of generated images rather than the ones produced by the latest generative networks. We trained the our model using ADAM \cite{Kingma2014} and set the learning rate to 0.0002 and momentum to 0.5. The batch size was set to 1. We implemented our network with the TensorFlow framework and trained it using NVIDIA TITAN X GPU. It took 15 hours to optimize our model.
\section{Experiments}
We compare the proposed model with several state-of-the-art methods: image-to-image transfer method (\ie, CycleGAN \cite{Zhu2017}), color constancy method (\ie, Gray Word (GW) \cite{Buchsbaum1980} ), image enhancement method (\ie, INT \cite{Gong2016}), and underwater image restoration methods (\ie, RED \cite{Galdran2015}, UWID \cite{Li2016} and UWIB \cite{Peng2017}). In our experiments, the test images were captured under varying underwater scenes. For fair comparisons, we did not compare our method with WaterGAN since WaterGAN was just available for underwater images taken under designated sites. We subjectively evaluate the visual quality of the results of different methods. Then, we conduct a user study since there is no metric designed for underwater image color correction available. Finally, we carry out application test on visual tasks in order to give additional evidence.

\subsection{Subjective assessment}
 \begin{figure*}[!tp]
\begin{center}
\begin{tabular}{c@{ }  c@{ } c@{ }  c@{ } c@{ }  c@{ } c@{ }  c@{ }}
\includegraphics[width=.115\textwidth]{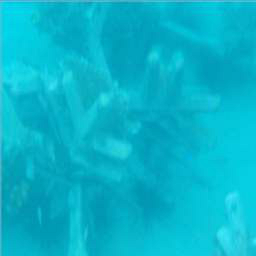}& \includegraphics[width=.115\textwidth]{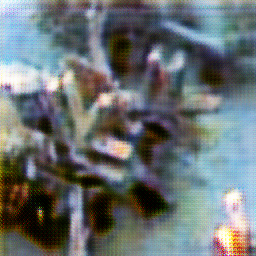}& \includegraphics[width=.115\textwidth]{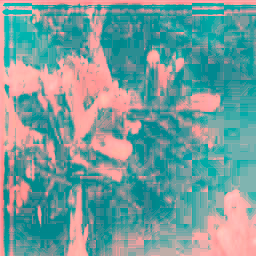}
 & \includegraphics[width=.115\textwidth]{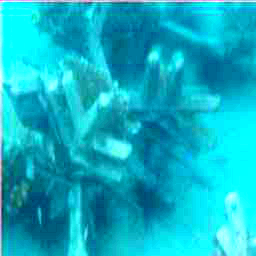}&
 \includegraphics[width=.115\textwidth]{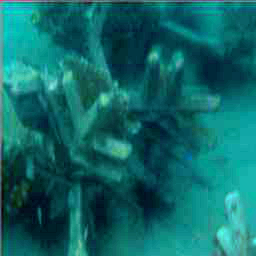}& \includegraphics[width=.115\textwidth]{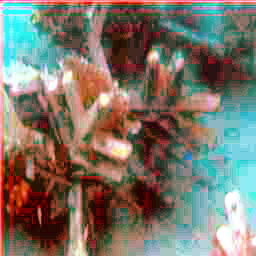} & \includegraphics[width=.115\textwidth]{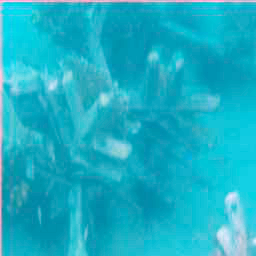}& \includegraphics[width=.115\textwidth]{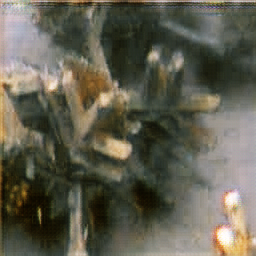}\\
\includegraphics[width=.115\textwidth]{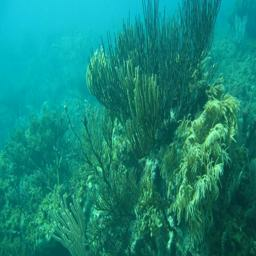}& \includegraphics[width=.115\textwidth]{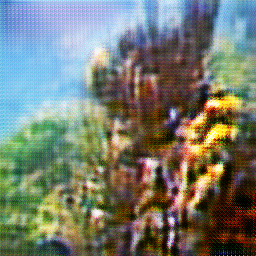}& \includegraphics[width=.115\textwidth]{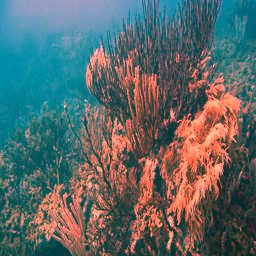}
 & \includegraphics[width=.115\textwidth]{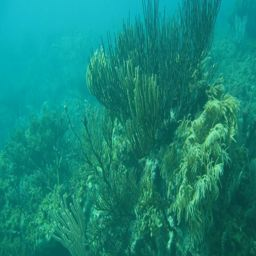}&
\includegraphics[width=.115\textwidth]{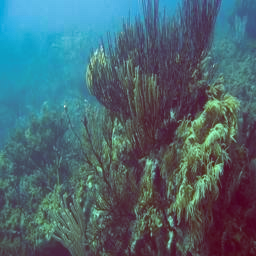}&
 \includegraphics[width=.115\textwidth]{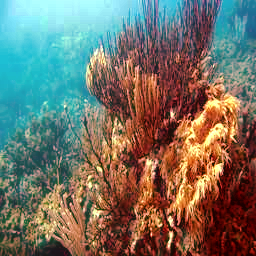} & \includegraphics[width=.115\textwidth]{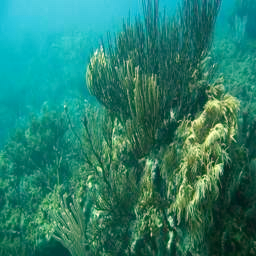}& \includegraphics[width=.115\textwidth]{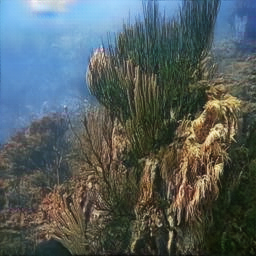}\\
\includegraphics[width=.115\textwidth]{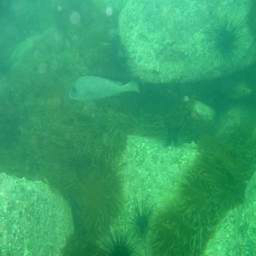}& \includegraphics[width=.115\textwidth]{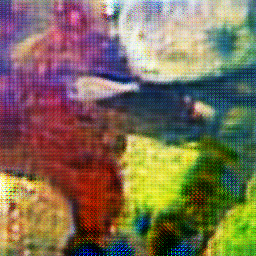}& \includegraphics[width=.115\textwidth]{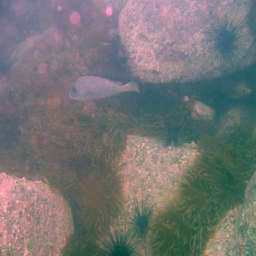}
 & \includegraphics[width=.115\textwidth]{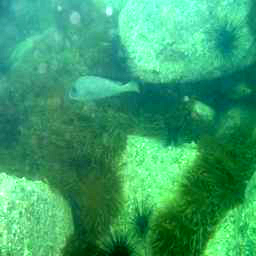}&
 \includegraphics[width=.115\textwidth]{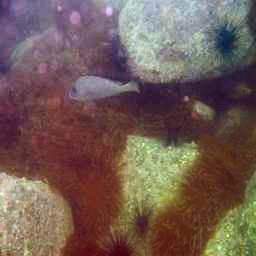}& \includegraphics[width=.115\textwidth]{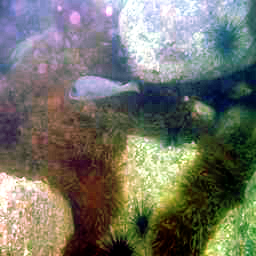} & \includegraphics[width=.115\textwidth]{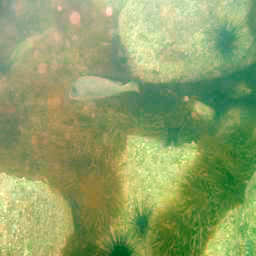}& \includegraphics[width=.115\textwidth]{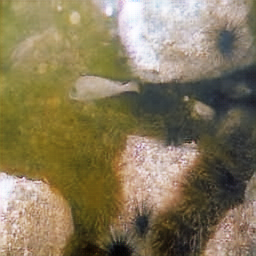}\\
\includegraphics[width=.115\textwidth]{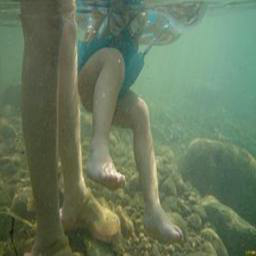}& \includegraphics[width=.115\textwidth]{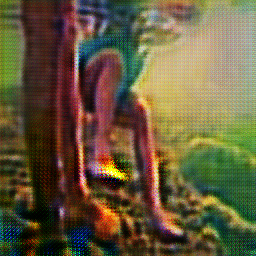}&  \includegraphics[width=.115\textwidth]{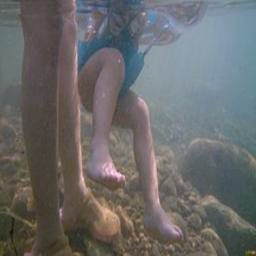}
 & \includegraphics[width=.115\textwidth]{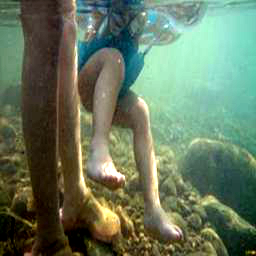}&
 \includegraphics[width=.115\textwidth]{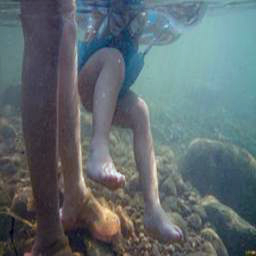}& \includegraphics[width=.115\textwidth]{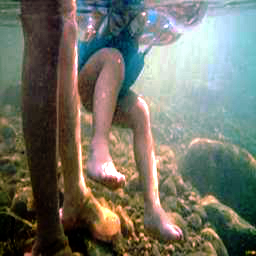} & \includegraphics[width=.115\textwidth]{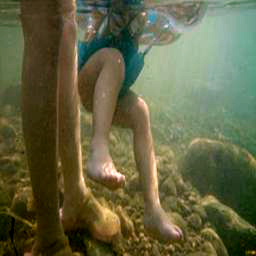}& \includegraphics[width=.115\textwidth]{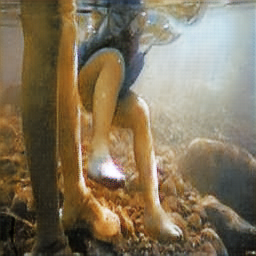}\\
\includegraphics[width=.115\textwidth]{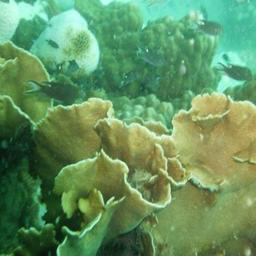}& \includegraphics[width=.115\textwidth]{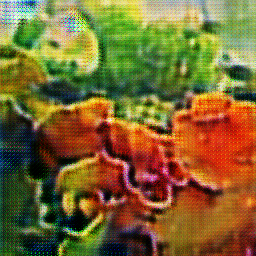}&  \includegraphics[width=.115\textwidth]{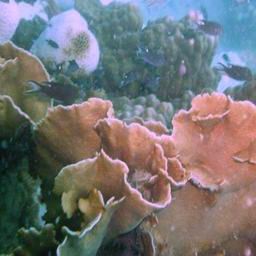}
  & \includegraphics[width=.115\textwidth]{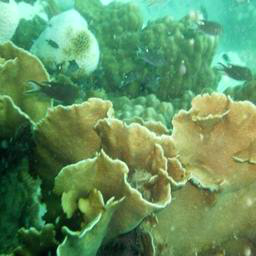}&
 \includegraphics[width=.115\textwidth]{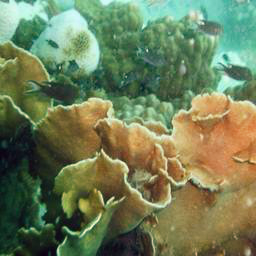}& \includegraphics[width=.115\textwidth]{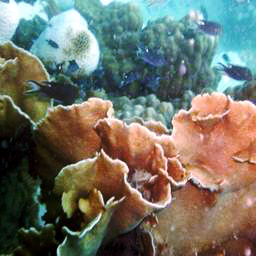} & \includegraphics[width=.115\textwidth]{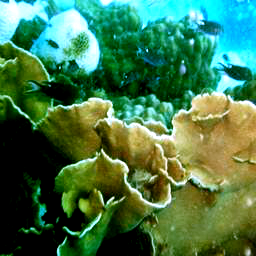}& \includegraphics[width=.115\textwidth]{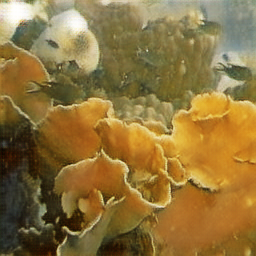}\\
 (a) Raws & (b) CycleGAN & (c) GW & (d) INT & (e) RED & (f) UWID & (g) UWIB & (h) Ours\\
\end{tabular}
\end{center}
\vspace{-2mm}
\caption{Visual quality comparisons on varying underwater scenes. From top to bottom are image I1-I5.}
\label{fig:quality assessment}
\vspace{-2mm}
\end{figure*}
In Figure~\ref{fig:quality assessment}, we show several results of different methods. In Figure~\ref{fig:quality assessment}, original CycleGAN produces blurring results because it tends to translate the content and structure of air image to the underwater image. GW method introduces color casts for some images when its assumption is not available. INT method just increases the brightness of underwater images because the prior learned from natural scenes is unavailable. Three underwater images restoration methods remove the haze effects and improve the contrast, however, the color is not well restored since the assumed optical parameters do not hold in some underwater scenes. On the contrary, our method can totally remove the greenish and bluish tone as if our results were taken without water, which leads to better visual quality.

\subsection{User study}

To perform a visual comparison in an objective way, we randomly selected 30 underwater images from our collected dataset for user study. The results of different methods were randomly displayed on the screen and compared with the corresponding raw underwater images. After that, we invited 10 participants who had experience with image processing to score results. There was no time limitation for each participant. Moreover, the participants did not know which results were produced by our method. The scores ranged from 1 (worst) to 8 (best). As baseline, we set the scores of raw underwater images to 3.  We expected that the good result has good contrast and visibility, abundant details, in especial the color as if the image was taken without water. On the contrary, the bad result has low visibility, over-enhanced regions, serious artifacts and noise, and inauthentic color. We first present in Table~\ref{table1} the average visual quality scores of the images shown in Figure~\ref{fig:quality assessment}.

\begin{table}[htbp]
\renewcommand{\arraystretch}{1}
\caption{ The average scores of the images presented in Figure~\ref{fig:quality assessment}}
\centering
\begin{tabular}{clccccc}
  \hline
 \textbf{Method} & \textbf{I1} & \textbf{I2} & \textbf{I3} & \textbf{I4} & \textbf{I5} \\
 \hline
Raws & 3 & 3 & 3& 3 & 3\\
CycleGAN  &  3.3 &  2.5 &  2.8 &  2.9 &  3.5 \\
GW  &  1.4 &  1.7 &  3.8 &  3.5 &  3 \\
INT & 3.1 & 3.1 & 3.5  & 4.9 & 3\\
RED   &   6.4 & 3.8  &  5.3 & 3.5  &  4.5\\
UWID      &  3.1 &   4.5  &  5.9 &   5.4  &  5.9\\
UWBI     &   3.5 & 4.1 & 5.5 &   4.5& 3.5 & \\
Ours     &   \textbf{7.5} &\textbf{ 7.5}  &\textbf{6.2} &  \textbf{6.2}  & \textbf{6.6}\\
\hline
\end{tabular}
\label{table1}
\vspace{\baselineskip}\end{table}

In Table~\ref{table1}, our results receives best scores, which indicates that, from a visual perspective, our method produces much better results. Additionally, the underwater image restoration methods also achieve good scores. The average visual quality scores for the selected 30 underwater images are 3, 2, 3, 3.3, 3.7, 4.6, 4 and 6.3 for Raws, CycleGAN, GW, INT, RED, UWID, UWBI, and Ours. More Results and scores are available in \url {https://li-chongyi.github.io/homepage.github.io/proj_Emerging_water.html}. Those results provide a realistic feedback that our method generates visually pleasing results.

\subsection{Application assessment}

\begin{figure}[tp]
\begin{center}
\begin{tabular}{c@{ }  c@{ } c@{ }  c@{ }   }
\includegraphics[width=.11\textwidth]{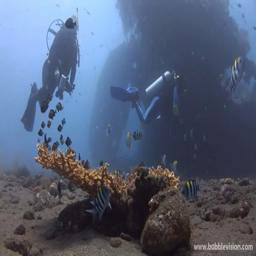} &
\includegraphics[width=.11\textwidth]{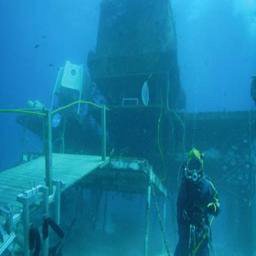}  &
\includegraphics[width=.11\textwidth]{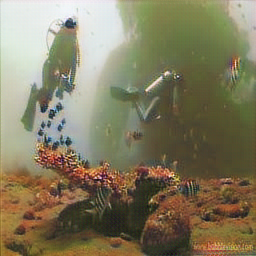} &
\includegraphics[width=.11\textwidth]{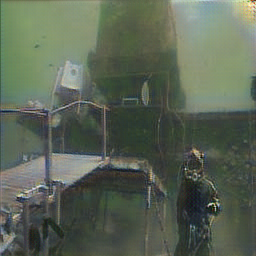}\\
(a) & (b)  & (c)  & (d) \\
\includegraphics[width=.11\textwidth]{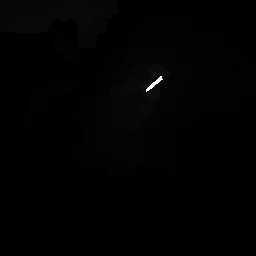}  &
\includegraphics[width=.11\textwidth]{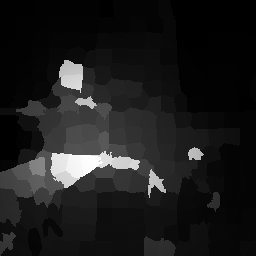}  &
\includegraphics[width=.11\textwidth]{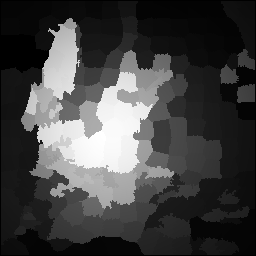}  &
\includegraphics[width=.11\textwidth]{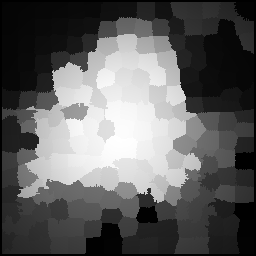}\\
(e) & (f)  & (g) & (h) \\
\end{tabular}
\end{center}
\vspace{-2mm}
\caption{Saliency detection test. (a) and (b) are raw underwater images. (c) and (d) are our results. (e) and (f) are the saliency maps of (a) and (b). (g) and (h) are the saliency maps of (c) and (d). }
\label{fig:saliency}
\vspace{-2mm}
\end{figure}

\begin{figure}[tp]
\begin{center}
\begin{tabular}{c@{ } c@{ }    }
\includegraphics[width=.22\textwidth]{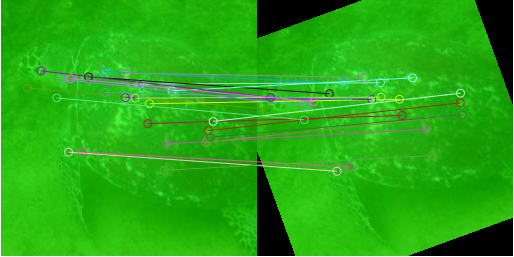}&
\includegraphics[width=.22\textwidth]{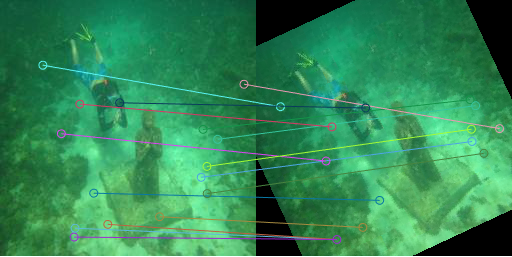}\\
(a) & (b) \\
\includegraphics[width=.22\textwidth]{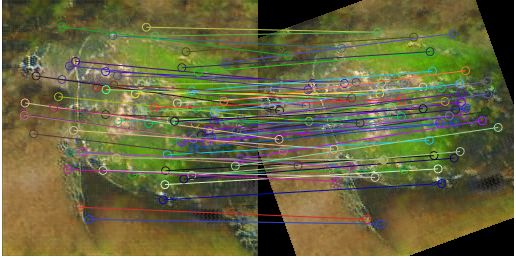} &
\includegraphics[width=.22\textwidth]{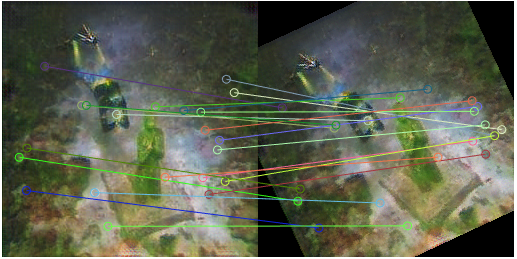}\\
 (c)  & (d) \\
\end{tabular}
\end{center}
\vspace{-2mm}
\caption{Keypoint matching test. (a) and (b) are keypoint matching maps of raw underwater images. (c) and (d) are keypoint matching maps of our results.}
\label{fig:surf}
\vspace{-2mm}
\end{figure}

To further demonstrate the effectiveness of the proposed methods, we carry out several application tests including saliency detection \cite{Peng2017TPAMI} and keypoint matching \cite{Bay2006}. In Figures~\ref{fig:saliency} and~\ref{fig:surf}, we present the results of application tests before and after using our method. For the limited space, we just present several examples.
After color corrected by our method, the results achieve better saliency detection performance and more matching points.  Application tests give additional evidence of the effectiveness of our method. 

\section{Conclusion}

In this letter, we presented a novel method for underwater image color correction. Based on the learned cross domain relations, the proposed method can remove color distortion by weakly supervised model. It is the first attempt that correcting the color casts of underwater image by weakly supervised learning. Furthermore, our method can be use as a guide for subsequent research of underwater image color correction. Experiments including quality evaluation, user study, and application assessment demonstrated the effectiveness of our method.

\end{document}